\documentclass[sigconf]{acmart}
\AtBeginDocument{%
  }

\usepackage{array}
\usepackage{amsmath}
\usepackage{booktabs}
\usepackage{multirow}
\usepackage{graphicx}
\usepackage[table]{xcolor}
\usepackage{balance}
\usepackage[normalem]{ulem}
\useunder{\uline}{\ul}{}

\copyrightyear{2026}
\acmYear{2026}

\setcopyright{cc}
\setcctype{by}

\acmConference[MM '26]
{Proceedings of the 34th ACM International Conference on Multimedia}
{November 10--14, 2026}
{Rio de Janeiro, Brazil}

\acmBooktitle{Proceedings of the 34th ACM International Conference on Multimedia
(MM '26), November 10--14, 2026, Rio de Janeiro, Brazil}

\acmISBN{979-8-4007-2213-4/2026/11}

\acmDOI{10.1145/3767308.3836595}

\begin{document}

%%
%% The "title" command has an optional parameter,
%% allowing the author to define a "short title" to be used in page headers.
\title{Beyond Surface Imitation: Contrastive Modeling for Reasoning Path Alignment in Multimodal In-Context Learning}

\author{Mingbo Yang}
\authornote{Mingbo Yang and Wenqiang Wang contributed equally to this work.}
\orcid{0009-0000-2952-5044}
\affiliation{%
  \institution{Sun Yat-Sen University}
  \city{Shenzhen}
  \state{Guangdong}
  \country{China}
}
\email{yangmb3@mail2.sysu.edu.cn}

\author{Wenqiang Wang}
\authornotemark[1]
\orcid{0009-0008-5034-1379}
\affiliation{%
  \institution{Sun Yat-Sen University}
  \city{Shenzhen}
  \state{Guangdong}
  \country{China}
}
\email{wangwq69@mail2.sysu.edu.cn}

\author{Zhaolu Kang}
\orcid{0009-0000-1163-1615}
\affiliation{%
  \institution{Peking University}
  \city{Beijing}
  \state{Beijing}
  \country{China}
}
\email{zlkang25@stu.pku.edu.cn}

\author{Peng Chen}
\orcid{0009-0001-0327-4053}
\affiliation{%
  \institution{Sun Yat-Sen University}
  \city{Shenzhen}
  \state{Guangdong}
  \country{China}
}
\email{chenpeng052@gmail.com}

\author{Yannan Chen}
\orcid{0009-0007-1896-3568}

\affiliation{%
  \institution{Sun Yat-Sen University}
  \city{Shenzhen}
  \state{Guangdong}
  \country{China}
}

\affiliation{%
  \institution{Pengcheng Laboratory}
  \city{Shenzhen}
  \state{Guangdong}
  \country{China}
}

\email{chenyn288@mail2.sysu.edu.cn}

\author{Sunshang Wang}
\orcid{0009-0004-9237-2887}
\affiliation{%
  \institution{Tianjin University of Science and Technology}
  \city{Tianjin}
  \state{Tianjin}
  \country{China}
}
\email{sunshangwww@mail.tust.edu.cn}

\author{Yan Xiao}
\authornote{Corresponding author.}
\orcid{0000-0002-2563-083X}
\affiliation{%
  \institution{Sun Yat-Sen University}
  \city{Shenzhen}
  \state{Guangdong}
  \country{China}
}
\email{xiaoyan.hhu@gmail.com}

\renewcommand{\shortauthors}{Mingbo Yang et al.}

%%
%% The abstract is a short summary of the work to be presented in the
%% article.
\begin{abstract}
In-context learning (ICL) is widely used in multimodal large language models (MLLMs) and achieves strong performance across a wide range of multimodal tasks. However, existing multimodal ICL methods often rely on surface level imitation of in-context demonstrations, making it difficult for MLLMs to align their responses with the reasoning path required by the given multimodal input. This limitation becomes more pronounced in complex multimodal tasks, thereby restricting further improvements in MLLM performance. To address this issue, we propose a new multimodal ICL framework that combines contrastive demonstration modeling with the self-refinement capability of MLLMs. Specifically, our framework reformulates each demonstration by explicitly contrasting a suboptimal response with a better response under the same input, together with a reasoning path that reveals how the response should be refined. This contrastive formulation makes the reasoning path toward the desired response more explicit and guides the MLLM beyond superficial imitation. Furthermore, because effective refinement depends on the current response, we introduce a response-conditioned retrieval mechanism to select demonstrations whose reasoning paths are more relevant to the current response. In addition, we use a lightweight alignment controller to predict response quality and determine whether further refinement is needed. Experiments on three types of multimodal tasks show that the proposed framework consistently improves MLLM performance, with particularly notable gains on visual question answering (VQA).
\end{abstract}


%%
%% Keywords. The author(s) should pick words that accurately describe
%% the work being presented. Separate the keywords with commas.
%% A "teaser" image appears between the author and affiliation
%% information and the body of the document, and typically spans the
%% page.
% \begin{teaserfigure}
%   \includegraphics[width=\textwidth]{sampleteaser}
%   \caption{Seattle Mariners at Spring Training, 2010.}
%   \Description{Enjoying the baseball game from the third-base
%   seats. Ichiro Suzuki preparing to bat.}
%   \label{fig:teaser}
% \end{teaserfigure}

% \received{20 February 2007}
% \received[revised]{12 March 2009}
% \received[accepted]{5 June 2009}

%%
%% This command processes the author and affiliation and title
%% information and builds the first part of the formatted document.
\maketitle

\section{Introduction}

\begin{figure*}[t]
    \centering
    \includegraphics[width=1\linewidth]{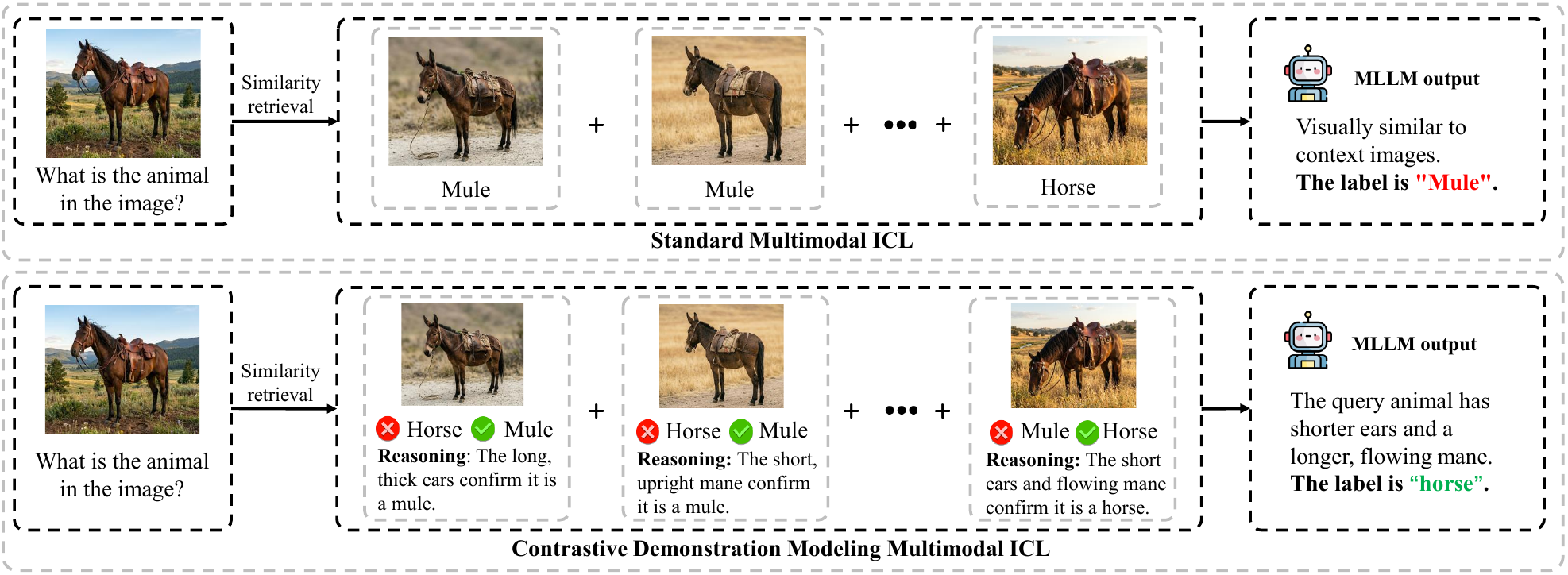}
    \caption{From surface level imitation to reasoning path alignment. Standard multimodal ICL mainly aligns the final output and may imitate surface patterns from visually similar demonstrations. In contrast, COMIL uses contrastive demonstrations to make explicit which visual evidence is relevant and how the current response should be revised, thereby guiding the MLLM toward the desired reasoning path.}
    \Description{Comparison between standard multimodal ICL and COMIL, illustrating how contrastive demonstrations and refinement guidance move the model from surface imitation toward evidence-grounded reasoning path alignment.}
\label{fig:motivation}
\end{figure*}

Multimodal In-Context Learning (multimodal ICL) has demonstrated remarkable capabilities in adapting Multimodal Large Language Models (MLLMs) to novel tasks~\citep{li2026catp,zhan2026retrieval,liao2026enhancing,li2025taco}. By conditioning MLLMs on demonstrations, multimodal ICL enables task adaptation at inference time without parameter updates~\citep{zheng2023can}. This makes multimodal ICL an attractive paradigm for multimodal tasks, as it flexibly leverages the pretrained knowledge of MLLMs~\citep{wang2024knowledgeable,yang2023supervised}.

Despite these advantages, current multimodal ICL paradigms often encourage surface level imitation rather than reasoning grounded in multimodal evidence~\citep{xu2025introspection,huang2025mimicking}. Here, surface level imitation refers to relying on superficial demonstration patterns without sufficiently grounding the response in query-relevant multimodal evidence. In standard multimodal ICL, MLLMs are conditioned on a sequence of input--output demonstrations~\citep{luo2024context,wang2024mixture}. However, when faced with complex scenarios that require fine-grained visual discrimination, MLLMs tend to rely on incorrect correlations. For instance, as shown in Figure~\ref{fig:motivation}, if the retrieved demonstrations share visually similar global features, the MLLM is prone to copying the most frequent textual labels from the context~\citep{fei2023mitigating}. By doing so, the MLLM bypasses the deeper visual analysis required for the task, matching only the output format while failing to ground its response in fine-grained visual evidence.

This issue reflects a fundamental limitation of standard demonstrations: they mainly align the final output, while overlooking the reasoning path that leads to the desired response~\citep{tang2025dawn,long2024does,wu2024beyond}. When an MLLM produces a response influenced by surface similarities, the key challenge is not merely to specify the target output, but to achieve reasoning path alignment by aligning the generated reasoning path with a latent evidence-grounded refinement relation under the multimodal input. In complex multimodal tasks, this requires explicit guidance on which visual evidence is truly relevant and how it should support revising the current response~\citep{sun2025exploring,zhou2024visual,zhang2023makes}. Standard demonstrations provide little such guidance, because they present only the target output rather than the reasoning path from a suboptimal response to a better one. Therefore, achieving reasoning path alignment requires demonstrations that make this refinement path explicit, enabling the MLLM to refine its response in a way that is better grounded in the underlying multimodal evidence.

To move beyond surface imitation, we propose COMIL, a novel framework that promotes reasoning path alignment in MLLMs through contrastive modeling of demonstrations and the self refinement capability of MLLMs~\citep{baldassini2024makes,doveh2024towards}. Rather than providing only a single target output, COMIL reformulates each standard ICL example under the same multimodal input as a structured demonstration tuple consisting of a suboptimal response, a better response, and a generated reasoning path describing how the response should be refined and what fine-grained visual evidence supports that refinement. By presenting both responses under the same input condition, this formulation makes explicit not only which response is better, but also how the current response should be improved and why that improvement is supported by the underlying multimodal evidence, thereby turning the demonstration into explicit guidance for reasoning path alignment.

Furthermore, because reasoning path alignment depends on the MLLM’s current response~\citep{guo2026problem,lee2025revise}, we couple these contrastive demonstrations with a retrieval mechanism guided by the current response. Rather than retrieving examples only from input similarity, COMIL selects demonstrations by additionally considering the similarity between their suboptimal responses and the MLLM's current response. In this way, the retrieved reasoning paths provide more relevant guidance for refining the current response. Combined with a lightweight alignment controller that predicts response quality and determines whether further refinement is needed, COMIL moves beyond passive output matching and instead guides generation toward more evidence-grounded refinement.

Extensive experiments across a diverse set of representative MLLMs verify the effectiveness of our framework. On three reasoning intensive tasks, COMIL shows clear improvements. For example, on VQAv2, COMIL improves the accuracy of Qwen3.5-9B to 81.9\%, achieving the best result among the compared methods. On Flickr30k, it improves the CIDEr score of Gemma-3-27B to 0.587. COMIL also remains sample efficient, maintaining favorable performance even with a limited number of retrieved demonstrations. In addition, the framework generalizes beyond open-source settings and achieves strong results on closed-source MLLMs. Overall, these results show that COMIL effectively improves multimodal ICL by reducing surface level imitation and promoting reasoning path alignment.

In summary, our main contributions are as follows:

\begin{itemize}
    \item \textbf{A New Framework for Reasoning Path Alignment in Multimodal ICL:} We propose COMIL, a new multimodal in-context learning framework that promotes reasoning path alignment by providing explicit guidance on which visual evidence is truly relevant and how it should support revising the current response.

    \item \textbf{Contrastive Demonstration Modeling with Response-Conditioned Retrieval and Alignment Control:} COMIL reformulates each standard ICL example as a structured contrastive demonstration tuple consisting of a suboptimal response, a better response, and the corresponding reasoning path. It further combines this formulation with response-conditioned retrieval and a lightweight alignment controller to guide the MLLM toward the desired reasoning path during inference.

    \item \textbf{Effective Results on Reasoning-Intensive Multimodal Tasks:} Extensive experiments across multiple MLLMs and multimodal datasets show that COMIL consistently improves performance, with particularly clear gains on reasoning intensive tasks. Additional reasoning-path evaluation and results on closed-source MLLMs further demonstrate the effectiveness and generality of the proposed framework.
\end{itemize}

\section{Related Work}

\subsection{Multimodal Large Language Models}

Multimodal Large Language Models (MLLMs) have recently achieved remarkable progress on a wide range of tasks, including visual question answering, image captioning, visual reasoning, and multimodal generation~\citep{erfani2026applications,yao2026towards,chen2026egoplan,jin2025efficient,kang2026hssbench,chen2026towards,chen2026dyc,chen2026domain}. By integrating visual encoders with large language models, together with large-scale vision-language pretraining, instruction tuning, and alignment, MLLMs have developed strong cross-modal understanding and generation capabilities, making them a key foundation for multimodal intelligence. Their ability to process visual and textual information within a unified generative framework enables them to perform diverse tasks with a single model and has significantly broadened the scope of vision-language applications.
Despite this progress, effective inference-time adaptation remains important for improving MLLM performance on complex multimodal tasks~\citep{xu2025learning,fan2025test}. In particular, how to better leverage the multimodal understanding and generation capabilities of MLLMs on complex tasks without parameter updates remains an important challenge~\citep{wu2026liquid,song2025bridge}. This has motivated growing interest in inference-time adaptation paradigms such as multimodal in-context learning.
\subsection{Multimodal In-Context Learning}

Multimodal In-Context Learning (multimodal ICL) has become an important paradigm for adapting MLLMs to downstream tasks~\citep{huang2026dissecting,li2026catp,zhou2026improving,li2025m,wang2026incomplete,wang2026task,wang2026dynamic}. By leveraging multimodal demonstrations composed of images, texts, and corresponding responses, multimodal ICL enables task adaptation at inference time without parameter updates~\citep{kumar2026act,gao2025aim}. This property makes multimodal ICL particularly attractive in practice, as it can flexibly adapt a pretrained MLLM to diverse tasks while avoiding the cost of additional training. More importantly, by presenting a small number of task-specific examples in context, multimodal ICL allows the MLLM to exploit its pretrained multimodal knowledge in a task-aware manner, making it a simple yet effective paradigm for multimodal classification, generation, and reasoning tasks. Existing studies have mainly improved multimodal ICL from the input side, such as demonstration retrieval, demonstration organization, prompt design, and instruction construction, showing that better demonstrations can substantially affect downstream performance~\citep{huang2026dissecting,chen2025can}.
Despite this progress, most existing methods still emphasize providing better demonstrations or improving alignment with the demonstrated target output. However, for complex multimodal tasks, recent observations suggest that MLLMs may still rely on superficial imitation of demonstrated responses rather than follow the reasoning path that leads to the correct result~\citep{nguyen2026iclr,huang2025mimicking}. This limitation motivates the need to go beyond better demonstration selection alone and to consider how demonstrations can more explicitly guide the MLLM’s reasoning.

\section{Method}

\begin{figure*}[t]
    \centering
    \includegraphics[width=1\linewidth]{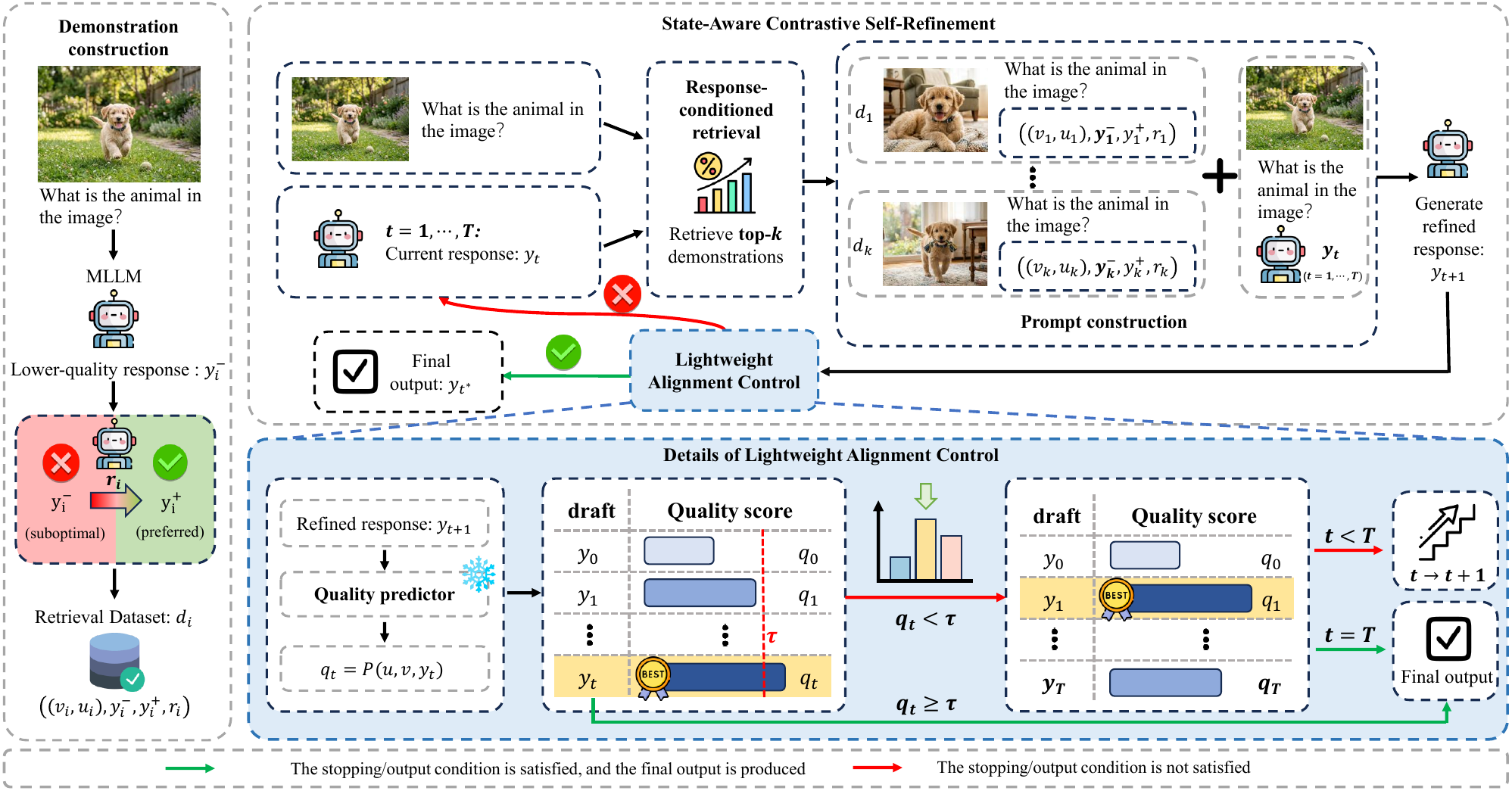}
    \caption{\textbf{Overall framework of COMIL.} Given an input $x=(v,u)$, the MLLM first generates an initial response. The retrieval dataset contains contrastive demonstrations $d_i=(x_i,y_i^{-},y_i^{+},r_i)$. At step $t$, the current response $y_t$ is used to retrieve the top-$k$ relevant demonstrations to generate $y_{t+1}$. The lightweight alignment controller predicts $q_t=\mathcal{P}(x,y_t)$ and compares it with $\tau$ to determine whether to continue refinement until $T$ or return the final response $y_{t^*}$.}
    \Description{Pipeline of COMIL showing the multimodal input, initial response generation, response-conditioned retrieval of contrastive demonstrations, iterative refinement, and quality-based stopping by the lightweight alignment controller.}
\label{fig:framework}
\end{figure*}

% \textcolor{red}{We formulate multimodal inference-time response improvement as a structured self-refinement framework.} Let the input \textcolor{blue}{be denoted} by $(v,u)$, where $v$ is the visual input, $u$ is the textual instruction or query, and $y$ is the textual response. Given $(v,u)$, the target MLLM defines a multimodal input--output mapping $p_\theta(y \mid v,u)$. Our framework improves this mapping at inference time through three coupled components: contrastive demonstration formulation, state-aware contrastive self-refinement, and lightweight quality control. Specifically, we reformulate demonstrations as low-to-high quality transitions, retrieve them conditioned on both the multimodal input and the current response, and iteratively refine the response under a lightweight quality controller. In this way, the framework makes the local correction structure of the multimodal mapping explicit and supports progressive response improvement during inference.
% \textcolor{blue}{why so many vocabulary like gpt's generation, input--output,state-aware}

\subsection{Problem Setup}\label{sec:problem_setup}

Let the multimodal input be denoted by $x=(v,u)$, where $v$ is the
visual input and $u$ is the textual instruction or query. Given $x$,
the target MLLM generates a textual response $y$ according to
$y \sim p_\theta(y \mid x)$.
In this work, reasoning path alignment refers to aligning the generated
reasoning path with a latent evidence-grounded refinement relation
under $x$: the path should diagnose missed or misused multimodal
evidence and support revising the current response toward a better one.
We define a latent alignment function $Q(x,y) \in \mathbb{R}$ to
conceptually score the resulting response state, where a larger value
indicates better alignment with this desired refinement relation.
Since $Q$ is not directly observable, it serves as a conceptual
objective, while its practical proxy is introduced in the lightweight
alignment control module. Under this formulation, the ideal target
response can be written conceptually as
$y^{*} = \arg\max_{y} Q(x,y)$.

To this end, COMIL is built on three coupled components: contrastive
demonstration modeling, response-conditioned retrieval, and lightweight
alignment control. Together, these components reformulate demonstrations
to make explicit the reasoning paths from suboptimal responses to better
responses, retrieve the demonstrations that are most relevant to the
current response, and guide the MLLM toward reasoning path alignment
during inference.

\subsection{Contrastive Demonstration Modeling}

To support reasoning path alignment, COMIL reformulates each demonstration
through contrastive modeling~\citep{baldassini2024makes,doveh2024towards}.
Rather than using a single target response, we explicitly show the MLLM
how a suboptimal response is refined into a better response together with
the corresponding reasoning path.
Formally, we represent the $i$th demonstration as
\begin{equation}
d_i = \bigl(x_i, y_i^{-}, y_i^{+}, r_i\bigr),
\end{equation}
where $x_i$ is the multimodal input, $y_i^{-}$ is a suboptimal response,
$y_i^{+}$ is a better response, and $r_i$ is a generated reasoning path
that describes how $y_i^{-}$ can be refined into $y_i^{+}$ under $x_i$.
Here, $r_i$ provides explicit refinement guidance rather than a direct
observation of the MLLM's internal reasoning process.

Under the fixed multimodal input $x_i$, $y_i^{-}$ and $y_i^{+}$
define a task-supervised refinement transition, while $r_i$ specifies
how the former can be revised toward the latter, thereby providing
explicit guidance for reasoning path alignment.
In the contrastive modeling, $y_i^{-}$ and $y_i^{+}$ form the contrasted
response pair. The reasoning path $r_i$ identifies the deficiency in
$y_i^{-}$, highlights the multimodal evidence relevant to the revision,
and shows how these cues support the transition to $y_i^{+}$. As a result, the contrastive demonstration provides explicit guidance
not only on the preferred response, but also on how the response should
be refined toward it. This supports reasoning path alignment at inference time.

\subsection{Response-Conditioned Retrieval and Refinement}

\paragraph{Construction of the Contrastive Retrieval Dataset.}
To support reasoning path alignment at inference time, we first
construct a retrieval dataset of contrastive demonstrations from a
selected subset of training examples. For each training instance $x_i$,
the same target MLLM used later at inference time is first prompted to
generate an initial response, which serves as the suboptimal response
$y_i^{-}$. The better response $y_i^{+}$ is taken directly from the
ground truth response in the dataset under the same multimodal input.
We then feed $x_i$, $y_i^{-}$, and $y_i^{+}$ into the same target MLLM
and prompt it to generate a reasoning path $r_i$, which describes how
$y_i^{-}$ can be refined into $y_i^{+}$. In this way, each training
instance is transformed into a contrastive demonstration
$d_i=(x_i,y_i^{-},y_i^{+},r_i)$, which explicitly represents a
generated refinement path from a suboptimal response state to a better
one~\citep{nguyen2026iclr,kang2025retrointext}. Repeating this process
over the selected training subset yields the retrieval dataset
$\mathcal{C}=\{d_i\}_{i=1}^{N}$ used in subsequent retrieval and
refinement.

\paragraph{Response-Conditioned Demonstration Retrieval.}
Given the retrieval dataset $\mathcal{C}=\{d_i\}_{i=1}^{N}$, the goal
of retrieval is to select contrastive demonstrations whose reasoning
paths are relevant to refining the current response. Since refinement
relevance depends on both the multimodal input and the current response,
retrieval is conditioned on $x$ and, whenever available, $y_t$~\citep{guo2026problem,lee2025revise}.
At the initial step, no response has been generated yet. We therefore
retrieve demonstrations only according to the multimodal input:
\begin{equation}
\mathcal{D}_0
=
\arg\max_{\mathcal{D}\subset \mathcal{C},\, |\mathcal{D}|=k}
\sum_{d_i\in\mathcal{D}}
\lambda_x\, s_x(x,x_i),
\,
y_0 = \mathcal{G}(x,\mathcal{D}_0),
\end{equation}
where $\mathcal{D}_0 \subset \mathcal{C}$ is the retrieved demonstration
set, $\mathcal{G}$ denotes the MLLM with the retrieved contrastive
demonstrations as context, and $s_x(x,x_i)$ measures the relevance
between the current multimodal input $x$ and the demonstration input
$x_i$.
Once the current response $y_t$ is available, retrieval is performed
according to a joint relevance score over the input and the current
response:
\begin{equation}
\begin{aligned}
\mathcal{D}_{t+1}
&=
\arg\max_{\mathcal{D}\subset \mathcal{C},\, |\mathcal{D}|=k}
\sum_{d_i\in\mathcal{D}}
\left[
\lambda_x\, s_x(x,x_i)
+
\lambda_y\, s_y(y_t,y_i^{-})
\right],\\
y_{t+1} &= \mathcal{G}(x,y_t,\mathcal{D}_{t+1}),
\end{aligned}
\end{equation}
where $s_y(y_t,y_i^{-})$ measures the similarity between the current
response $y_t$ and the suboptimal response $y_i^{-}$ in the
demonstration, and $\lambda_x,\lambda_y \ge 0$ balance the contributions
of input relevance and response similarity.
We use response similarity as a practical proxy for refinement
relevance. When $y_t$ is similar to $y_i^{-}$, the associated reasoning
path $r_i$ is more likely to describe a correction pattern relevant to
the current response. The retrieved demonstration therefore provides
not only a better response but also potentially useful guidance for
revision. Response similarity does not guarantee the same error mode,
but enables retrieval to incorporate the current response state in
addition to input similarity.

\paragraph{Refinement with Retrieved Demonstrations.}
Under this retrieval mechanism, iterative refinement aims to move the
current response toward stronger reasoning path alignment, rather than
guaranteeing monotonic improvement at every step~\citep{xu2025toward,chen2025towards,ke2025survey}. In practice, the
target MLLM uses the retrieved contrastive demonstrations to revise the
current response according to refinement patterns relevant to its
current state. In this way, retrieval and refinement are tightly
coupled: retrieval identifies relevant refinement guidance, while the
MLLM uses this guidance to revise the current response toward the
evidence-grounded refinement relation defined in
Section~\ref{sec:problem_setup}.

\begin{figure}[t]
    \centering
    \includegraphics[width=1\linewidth]{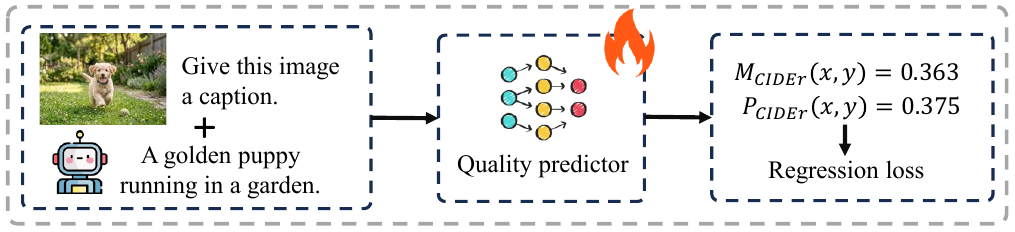}
    \caption{\textbf{Training pipeline of the lightweight quality predictor.} The predictor takes an input and a candidate response as input, predicts the response evaluation metric, and is trained with a regression loss against the actual metric $M(x,y)$.}
    \Description{Training pipeline in which the lightweight predictor receives the multimodal input and a candidate response, predicts response quality, and is optimized with a regression loss against the observed task metric.}
\label{fig:controller_training}
\end{figure}

\subsection{Lightweight Alignment Control}

While reasoning path alignment is the target of refinement, it is not
directly observable at inference time. We therefore introduce a
lightweight alignment controller that predicts response quality as a
proxy for controlling the refinement trajectory. During controller
training, the task-specific evaluation metric provides an observable
supervision signal: under a fixed multimodal input $x$, a response that
better follows the desired evidence-grounded refinement relation is more
likely to match the expected output and achieve a better task metric~\citep{madaan2023self,paul2024refiner}. We therefore use the
response-level evaluation metric to supervise the controller.

Formally, let $M(x,y)$ denote the actual evaluation metric of response
$y$ under input $x$, where a larger value indicates better task
performance. As defined in Section~\ref{sec:problem_setup}, $Q(x,y)$ is a latent
alignment property and cannot be directly observed. We therefore use
$M(x,y)$ as a practical supervision signal and train a lightweight
predictor $\mathcal{P}$ to estimate the quality of each intermediate
response, $q_t = \mathcal{P}(x,y_t)$, where $q_t$ is the predicted
quality score at refinement step $t$.
The predictor is trained with supervision derived from the actual metric
$M(x,y_t)$; thus, $q_t$ predicts response quality rather than directly
measuring the latent alignment function $Q$. For ease of exposition,
we refer to this predictor as the \emph{lightweight alignment controller}.
The relation between task performance and reasoning-path alignment is
further examined in Section~\ref{sec:rpa_analysis}.
This alignment control is helpful because iterative refinement does not
always improve the response monotonically. Although the retrieved
demonstrations are selected to support reasoning path alignment, later
refinement steps may still preserve, weaken, or distort the desired
refinement direction~\citep{huang2023crossgnn}. The controller therefore
evaluates the predicted quality of each intermediate response and
prevents the final decision from depending only on the latest step. In
this way, it helps retain a higher-quality response among the explored
candidates.

Based on the predicted quality score, we introduce an acceptance
threshold $\tau$ for early stopping. In practice, $\tau$ is determined
from the training data used for the alignment controller: we set $\tau$
to the evaluation metric corresponding to the top 25\% of the training
responses, which are ranked by the metric in descending order. This
choice encourages early stopping on high-quality responses while
avoiding an overly strict threshold that would make successful stopping
difficult. If $q_t \ge \tau$, the current response is regarded as
sufficiently high-quality and refinement stops. Otherwise, refinement
continues until the maximum refinement budget $T$ is reached. To unify
early stopping and final selection, we define
\begin{equation}
t^* =
\begin{cases}
\min \{ t \mid q_t \ge \tau \}, & \text{if } \exists\, t \in \{0,\ldots,T\},\, q_t \ge \tau,\\
\arg\max_{t \in \{0,\ldots,T\}} q_t, & \text{otherwise},
\end{cases}
\end{equation}
and take $y_{t^*}$ as the final response.

Through this design, the controller neither generates reasoning paths
nor directly measures reasoning-path alignment. Instead, it uses the
predicted response quality as a practical proxy to determine when
refinement should stop or which intermediate response should be
selected. Together with contrastive demonstration modeling and response-conditioned retrieval, it stabilizes the inference-time refinement
process toward reasoning path alignment.

\section{Experiment}

% Please add the following required packages to your document preamble:
% \usepackage{booktabs}
% \usepackage{multirow}
% \usepackage{graphicx}
% \usepackage[normalem]{ulem}
% \useunder{\uline}{\ul}{}
\begin{table*}[t]
\centering
\caption{Main results on four multimodal datasets and four open-source MLLMs, including recent multimodal ICL baselines. Best and second-best results are highlighted in bold and underlined, respectively; ``--'' denotes unavailable results.}
\label{tab:main_exp}

\resizebox{\textwidth}{!}{%
\begin{tabular}{@{}cc|cccc|cccc@{}}
\toprule

\multicolumn{1}{c|}{\multirow{3}{*}{\textbf{Category}}}
& \textbf{MLLMs}
& \textbf{Gemma-3-27B}
& \textbf{InternVL3.5-14B}
& \textbf{Qwen3.5-9B}
& \textbf{Qwen3.5-4B}
& \textbf{Gemma-3-27B}
& \textbf{InternVL3.5-14B}
& \textbf{Qwen3.5-9B}
& \textbf{Qwen3.5-4B}
\\ \cmidrule(l){2-10}

\multicolumn{1}{c|}{}
& \textbf{Metrics}
& \multicolumn{4}{c|}{Accuracy (\%) $\uparrow$}
& \multicolumn{4}{c}{CIDEr $\uparrow$}
\\ \cmidrule(l){2-10}

\multicolumn{1}{c|}{}
& \textbf{Method}
& \multicolumn{4}{c|}{Image Classification: CIFAR10}
& \multicolumn{4}{c}{Image Captioning: Flickr30k}
\\ \midrule \midrule

\multicolumn{1}{c|}{\multirow{3}{*}{\begin{tabular}[c]{@{}c@{}}Retrieval\\based\\ICL\end{tabular}}}
& CLIPRE
& 95.7 & 92.3 & 92.5 & 91.1
& 0.297 & 0.287 & 0.316 & 0.294 \\

\multicolumn{1}{c|}{}
& KNN
& 95.3 & 95.0 & 93.7 & 92.1
& 0.346 & 0.352 & 0.410 & 0.387 \\

\multicolumn{1}{c|}{}
& CR
& 93.6 & 91.8 & 88.4 & 88.5
& 0.303 & 0.381 & 0.388 & 0.369
\\ \midrule

\multicolumn{1}{c|}{\multirow{2}{*}{\begin{tabular}[c]{@{}c@{}}Train based\\ICL\end{tabular}}}
& LCL
& {\ul 96.5} & 94.3 & {\ul 96.4} & 91.4
& {\ul 0.514} & 0.384 & {\ul 0.459} & 0.340 \\

\multicolumn{1}{c|}{}
& MimIC
& 91.4 & {\ul 95.6} & 94.6 & 95.5
& 0.349 & 0.378 & 0.429 & 0.394
\\ \midrule

\multicolumn{1}{c|}{\multirow{2}{*}{CoT}}
& Few-shot CoT
& 93.6 & 95.4 & 91.8 & 94.2
& 0.390 & 0.331 & 0.390 & 0.402 \\

\multicolumn{1}{c|}{}
& Self-Consistency CoT
& 94.0 & 95.4 & 93.7 & \textbf{99.0}
& 0.377 & 0.322 & 0.362 & 0.342
\\ \midrule

\multicolumn{1}{c|}{\multirow{3}{*}{Self-Refine}}
& Iteration
& 87.3 & 83.3 & 83.9 & 85.8
& 0.169 & 0.131 & 0.075 & 0.098 \\

\multicolumn{1}{c|}{}
& Self-Refine
& 89.6 & 83.2 & 84.0 & 85.0
& 0.114 & 0.111 & 0.060 & 0.014 \\

\multicolumn{1}{c|}{}
& SC-Captioner
& 91.6 & 91.9 & 93.9 & 94.4
& 0.472 & 0.393 & 0.434 & 0.425
\\ \midrule

\multicolumn{1}{c|}{\multirow{3}{*}{\begin{tabular}[c]{@{}c@{}}Recent\\Multimodal\\ICL\end{tabular}}}
& AIM
& -- & 94.2 & 95.4 & 91.9
& -- & {\ul 0.431} & 0.445 & 0.448 \\

\multicolumn{1}{c|}{}
& TACO
& -- & 91.0 & 91.0 & 86.6
& -- & 0.399 & 0.373 & 0.420 \\

\multicolumn{1}{c|}{}
& M$^2$IV
& -- & 95.5 & 93.7 & 94.6
& -- & 0.412 & 0.428 & {\ul 0.466}
\\ \midrule

\multicolumn{2}{c|}{\textbf{COMIL (Ours)}}
& \textbf{97.4}
& \textbf{98.4}
& \textbf{97.4}
& {\ul 96.5}
& \textbf{0.587}
& \textbf{0.517}
& \textbf{0.499}
& \textbf{0.493}
\\ \midrule

\multicolumn{1}{c|}{\multirow{2}{*}{\textbf{Category}}}
& \textbf{Metrics}
& \multicolumn{4}{c|}{Accuracy (\%) $\uparrow$}
& \multicolumn{4}{c}{Accuracy (\%) $\uparrow$}
\\ \cmidrule(l){2-10}

\multicolumn{1}{c|}{}
& \textbf{Method}
& \multicolumn{4}{c|}{Visual Question Answering: VQAv2}
& \multicolumn{4}{c}{Visual Question Answering: OKVQA}
\\ \midrule \midrule

\multicolumn{1}{c|}{\multirow{3}{*}{\begin{tabular}[c]{@{}c@{}}Retrieval\\based\\ICL\end{tabular}}}
& CLIPRE
& 57.0 & 74.4 & 61.8 & 69.2
& 31.2 & 42.0 & 33.6 & 38.7 \\

\multicolumn{1}{c|}{}
& KNN
& 67.3 & 78.1 & 66.5 & 64.9
& 51.7 & 52.2 & {\ul 53.9} & 47.6 \\

\multicolumn{1}{c|}{}
& CR
& 65.3 & 78.0 & 71.2 & 69.7
& 51.4 & 50.3 & 49.2 & 43.6
\\ \midrule

\multicolumn{1}{c|}{\multirow{2}{*}{\begin{tabular}[c]{@{}c@{}}Train based\\ICL\end{tabular}}}
& LCL
& {\ul 70.2} & 79.7 & 69.6 & 77.3
& 52.3 & 52.7 & 41.5 & 42.0 \\

\multicolumn{1}{c|}{}
& MimIC
& 62.9 & 72.7 & 68.7 & 68.4
& 42.1 & 55.0 & 35.3 & 54.9
\\ \midrule

\multicolumn{1}{c|}{\multirow{2}{*}{CoT}}
& Few-shot CoT
& 64.5 & 79.8 & 61.4 & 59.2
& 53.6 & 56.3 & 46.3 & 48.7 \\

\multicolumn{1}{c|}{}
& Self-Consistency CoT
& 66.0 & \textbf{81.1} & 62.9 & 60.5
& {\ul 53.7} & 57.2 & 51.1 & 54.7
\\ \midrule

\multicolumn{1}{c|}{\multirow{3}{*}{Self-Refine}}
& Iteration
& 31.1 & 25.1 & 20.6 & 30.1
& 32.4 & 22.3 & 27.5 & 17.3 \\

\multicolumn{1}{c|}{}
& Self-Refine
& 21.4 & 26.6 & 26.6 & 33.4
& 42.4 & 38.5 & 44.8 & 42.6 \\

\multicolumn{1}{c|}{}
& SC-Captioner
& 65.0 & 66.1 & 63.8 & 61.4
& 43.7 & 49.4 & 29.2 & 39.1
\\ \midrule

\multicolumn{1}{c|}{\multirow{3}{*}{\begin{tabular}[c]{@{}c@{}}Recent\\Multimodal\\ICL\end{tabular}}}
& AIM
& -- & 77.2 & 80.2 & 77.8
& -- & {\ul 59.8} & 51.0 & \textbf{62.5} \\

\multicolumn{1}{c|}{}
& TACO
& -- & 75.1 & {\ul 80.9} & {\ul 78.7}
& -- & {\ul 59.8} & 50.1 & 59.2 \\

\multicolumn{1}{c|}{}
& M$^2$IV
& -- & 73.6 & 79.9 & 76.9
& -- & 51.1 & 53.1 & 51.7
\\ \midrule

\multicolumn{2}{c|}{\textbf{COMIL (Ours)}}
& \textbf{74.0}
& {\ul 81.0}
& \textbf{81.9}
& \textbf{81.2}
& \textbf{54.7}
& \textbf{61.7}
& \textbf{56.8}
& {\ul 61.2}
\\ \bottomrule

\end{tabular}%
}
\end{table*}

\subsection{Experimental Setup}

\paragraph{MLLMs and Datasets.}
We evaluate COMIL on four representative open-source multimodal large language models (MLLMs), including Gemma-3-27B~\citep{gemma_2025}, InternVL3.5-14B~\citep{wang2025internvl3_5}, Qwen3.5-9B~\citep{qwen3.5}, and Qwen3.5-4B~\citep{qwen3.5}. 
Following the main experimental setting, we conduct experiments on four datasets: CIFAR10~\citep{krizhevsky2009learning} for image classification, Flickr30k~\citep{young2014image} for image captioning, and VQAv2~\citep{goyal2017making} and OKVQA~\citep{schwenk2022okvqa} for visual question answering. 

\paragraph{Tasks and Metrics.}
We evaluate COMIL on three types of multimodal tasks with task-specific metrics: image classification, image captioning, and visual question answering. These tasks cover multimodal classification, generation, and reasoning, enabling evaluation across diverse multimodal settings. For image classification, we report Accuracy. For image captioning, we report CIDEr~\citep{vedantam2015cider}. For visual question answering, we report VQA Accuracy~\citep{antol2015vqa}. For all metrics, higher values indicate better performance.

\paragraph{Baselines.}
We compare COMIL with representative baselines from five categories, as shown in Table~\ref{tab:main_exp}. Specifically, we include \textbf{Retrieval-based ICL} methods, including CLIPRE~\citep{radford2021learning}, KNN~\citep{guo2003knn}, and CR~\citep{liu2004cluster}; \textbf{Train-based ICL} methods, including LCL~\citep{tai2024link} and MimIC~\citep{jiang2025mimic}; \textbf{CoT} methods, including Few-shot CoT~\citep{kim2023cot} and Self-Consistency CoT~\citep{wang2022self}; \textbf{Self-Refine} methods, including Iteration, Self-Refine~\citep{madaan2023self}, and SC-Captioner~\citep{zhang2025sc}; and recent \textbf{Multimodal ICL} methods, including AIM~\citep{gao2025aim}, TACO~\citep{li2025taco}, and M$^2$IV~\citep{li2025m}. These baselines cover retrieval-based demonstration selection, training-enhanced in-context learning, reasoning-oriented prompting, iterative refinement, and recent multimodal ICL approaches.

% \paragraph{Inference Details.}
% For each test instance, COMIL generates an initial response and performs up to 3 refinement iterations using top-$k=10$ retrieved contrastive demonstrations.
% In our implementation, the lightweight alignment controller uses openai/clip-vit-base-patch32~\citep{radford2021learning} as the image encoder and google-bert/bert-base-uncased~\citep{devlin2019bert} as the text encoder. 
% For controller training, we use part of the training split of each dataset. The retrieval set is constructed from the last 500 samples of the corresponding training split and is kept disjoint from the controller training set. The main experiments are conducted on the test split of each dataset. Therefore, the controller training set, retrieval set, and evaluation set are mutually exclusive, and no data instance is reused across them. 
% We use the same experimental protocol across different target MLLMs and datasets, and compare all methods under consistent task settings for fair evaluation. All experiments are implemented in vLLM and conducted on NVIDIA A100 GPUs. Specifically, InternVL3.5-14B, Qwen3.5-9B, and Qwen3.5-4B are run on a single NVIDIA A100 GPU, while Gemma-3-27B is run on two NVIDIA A100 GPUs.
\paragraph{Inference Details.}
For each test instance, COMIL generates an initial response and performs up to 3 refinement iterations using top-$k=10$ retrieved contrastive demonstrations.
In our implementation, the lightweight alignment controller uses openai/clip-vit-base-patch32~\citep{radford2021learning} as the image encoder and google-bert/bert-base-uncased~\citep{devlin2019bert} as the text encoder.
For response-conditioned retrieval, both input and response similarities are computed using cosine similarity, with the two retrieval weights set to 0.5.
The initial response and each refinement round use a maximum generation length of 2048 tokens.
Both controller encoders are frozen, and their representations are fused and scored by a three-layer MLP with a hidden dimension of 512 and dropout of 0.1.
For controller training, we use part of the training split of each dataset. The retrieval set is constructed from the last 500 samples of the corresponding training split and is kept disjoint from the controller training set. The main experiments are conducted on the test split of each dataset. Therefore, the controller training set, retrieval set, and evaluation set are mutually exclusive, and no data instance is reused across them.
We use the same experimental protocol across different target MLLMs and datasets, and compare all methods under consistent task settings for fair evaluation. 

\subsection{Main Results}

Table~\ref{tab:main_exp} reports the main results across four multimodal datasets and four open-source MLLMs, including comparisons with recent multimodal ICL methods. Overall, COMIL achieves the best performance in 13 out of 16 MLLM--dataset settings and ranks second in the remaining three, demonstrating strong and consistent performance across multimodal classification, captioning, and visual question answering.
On CIFAR-10, COMIL achieves the best result on three of the four MLLMs and ranks second on Qwen3.5-4B. Its advantages are more consistent on multimodal generation and reasoning tasks. On Flickr30k, COMIL achieves the best performance under all four MLLMs; for example, it reaches 0.517 CIDEr on InternVL3.5-14B, outperforming the strongest baseline AIM by 0.086. On VQAv2, COMIL achieves the best result on three MLLMs and ranks second on InternVL3.5-14B. In particular, it reaches 81.9\% accuracy on Qwen3.5-9B, exceeding TACO by 1.0 percentage point. On OKVQA, COMIL again achieves the best performance in three of the four settings and ranks second on Qwen3.5-4B.
These results show that COMIL remains competitive not only against conventional retrieval, CoT, and self-refinement baselines, but also against recent multimodal ICL methods. Another notable finding is that simple Self-Refine based baselines perform poorly across most multimodal tasks. This suggests that the gains of COMIL do not come from iterative refinement alone, but from the combination of contrastive demonstration modeling, response-conditioned retrieval, and lightweight alignment control.

\subsection{Ablation Study}

\begin{table}[t]
\centering
\caption{Ablation study of contrastive demonstration modeling in COMIL\@. As more components of the contrastive four-tuple are introduced, performance improves consistently across datasets and target MLLMs.}
\label{tab:ablation1}
\resizebox{0.48\textwidth}{!}{%
\begin{tabular}{@{}cccc|cc|cc@{}}
\toprule
\multicolumn{4}{c|}{\textbf{MLLMs}}                   & InternVL3.5-14B  & Qwen3.5-4B  & InternVL3.5-14B   & Qwen3.5-4B   \\ \midrule
\multicolumn{4}{c|}{\textbf{Datasets}}                 & \multicolumn{2}{c|}{Flickr30k} & \multicolumn{2}{c}{VQAv2}        \\ \midrule
$(v_i,u_i)$ & $y_i^{-}$ & $y_i^{+}$ & $r_i$ & \multicolumn{2}{c|}{CIDEr$\uparrow$}     & \multicolumn{2}{c}{Accuracy(\%)$\uparrow$} \\ \midrule 
$\checkmark$           &      $\times$     & $\checkmark$         &    $\times$    & 0.387            & 0.347       & 70.9              & 69.0         \\
$\checkmark$           & $\checkmark$         & $\checkmark$         &   $\times$     & 0.475            & 0.358       & 77.1              & 71.2         \\
$\checkmark$           & $\checkmark$         & $\checkmark$         & $\checkmark$      & 0.517            & 0.493       & 81.0              & 81.2         \\ \bottomrule
\end{tabular}
}
\end{table}

\begin{table}[t]
\centering
\caption{Ablation study of response-conditioned retrieval in COMIL\@. Response-conditioned retrieval consistently outperforms random retrieval across datasets and target MLLMs.}
\label{tab:ablation2}
\resizebox{0.48\textwidth}{!}{%
\begin{tabular}{@{}c|cc|cc@{}}
\toprule
\textbf{MLLMs}                    & InternVL3.5-14B     & Qwen3.5-4B     & InternVL3.5-14B        & Qwen3.5-4B        \\ \midrule
\textbf{Datasets}                 & \multicolumn{2}{c|}{Flickr30k}       & \multicolumn{2}{c}{VQAv2}                  \\ \midrule
\textbf{Retrieval}          & \multicolumn{2}{c|}{CIDEr$\uparrow$} & \multicolumn{2}{c}{Accuracy(\%)$\uparrow$} \\ \midrule 
Random                       & 0.431               & 0.341          & 75.8                 & 69.9            \\
Response-conditioned & 0.517               & 0.493          & 81.0                 & 81.2            \\ \bottomrule
\end{tabular}
}
\end{table}

\begin{table}[t]
\centering
\caption{Ablation study of the lightweight alignment control module in COMIL\@. The alignment controller consistently improves final performance.}
\label{tab:ablation3}
\resizebox{0.48\textwidth}{!}{%
\begin{tabular}{@{}c|cc|cc@{}}
\toprule
\textbf{MLLMs}            & InternVL3.5-14B     & Qwen3.5-4B     & InternVL3.5-14B        & Qwen3.5-4B        \\ \midrule
\textbf{Datasets}          & \multicolumn{2}{c|}{Flickr30k}       & \multicolumn{2}{c}{VQAv2}                  \\ \midrule
\textbf{Alignment Control}  & \multicolumn{2}{c|}{CIDEr$\uparrow$} & \multicolumn{2}{c}{Accuracy(\%)$\uparrow$} \\ \midrule
$\times$ & 0.491               & 0.407          & 76.3                   & 75.2              \\
$\checkmark$ & 0.517               & 0.493          & 81.0                   & 81.2              \\ \bottomrule
\end{tabular}
}
\end{table}

We further analyze how each component contributes to reasoning path alignment during inference. Following the design of COMIL, we study three aspects: contrastive demonstration modeling, response-conditioned retrieval, and lightweight alignment control.

\paragraph{Effect of Contrastive Demonstration Modeling}

Table~\ref{tab:ablation1} evaluates the effect of the contrastive four-tuple formulation. When the demonstrations contain only the multimodal input $(v_i,u_i)$ and the better response $y_i^{+}$, performance drops substantially on both datasets and both target MLLMs. Adding the suboptimal response $y_i^{-}$ but removing the reasoning path $r_i$ results in a less severe performance drop. A standard positive-only demonstration mainly shows the MLLM a preferred response, but does not specify how the current response should be refined under the same input condition. In contrast, introducing $y_i^{-}$ makes the contrast between suboptimal and better responses explicit, while adding $r_i$ further clarifies how the response should be refined and what multimodal evidence supports that refinement. This enables the demonstrations to provide more informative guidance for reasoning path alignment.

\paragraph{Effect of Response-Conditioned Retrieval}

Table~\ref{tab:ablation2} studies the retrieval strategy. Replacing response-conditioned retrieval with random retrieval consistently degrades performance across both tasks and both MLLMs. This result shows that the effectiveness of retrieval in COMIL depends not simply on the availability of demonstrations, but on whether the retrieved demonstrations provide reasoning paths relevant to the current response under the same input condition. By retrieving such demonstrations, response-conditioned retrieval offers more effective guidance for refining the current response toward the desired reasoning path.

\paragraph{Effect of Lightweight Alignment Control}

Table~\ref{tab:ablation3} examines the lightweight alignment controller. Removing alignment control causes a consistent but smaller performance drop than removing the contrastive formulation or the retrieval mechanism. This shows that alignment control is not the main source of gains, but it still contributes to the final performance. Its role is better understood as stabilizing refinement rather than providing reasoning guidance by itself. Contrastive demonstration modeling and response-conditioned retrieval determine how the current response should be refined, while the controller predicts the quality of each intermediate response and determines when refinement should stop or which response should be selected. In this way, the controller helps avoid unnecessary or harmful refinement steps and makes the overall refinement process more reliable across datasets and MLLMs.

\section{Discussion}
% Please add the following required packages to your document preamble:
% \usepackage{booktabs}
\subsection{Direct Evaluation of Reasoning Path Alignment}
\label{sec:rpa_analysis}

Final task metrics do not directly measure whether the
refinement paths are grounded in query-relevant multimodal evidence.
We therefore use GPT-5.5 to evaluate each reasoning path along four
0--4 criteria: evidence mention, evidence correctness, task relevance,
and response--evidence consistency. The aggregated score is normalized
to $[0,1]$ as the reasoning path alignment (RPA) score. We also report
the sample-level Spearman correlation $\rho$ between RPA and the task metric.
As shown in Table~\ref{tab:rpa_analysis}, COMIL achieves the highest RPA
score in five of the six settings. Moreover, its Spearman $\rho$
remains consistently positive, ranging from 0.55 to 0.68 with an
average of 0.63. These results indicate that the task-performance
gains of COMIL are generally accompanied by refinement paths that
are better grounded in multimodal evidence under the adopted
evaluation protocol. We note that RPA evaluates the generated
refinement path and should not be interpreted as a direct measure
of the model's unobservable internal reasoning process.

\begin{table}[t]
\centering
\caption{Direct reasoning-path alignment evaluation. RPA denotes the normalized reasoning-path alignment score, and $\rho$ denotes the sample-level Spearman correlation with the task metric.}
\label{tab:rpa_analysis}
\resizebox{\columnwidth}{!}{%
\begin{tabular}{@{}c|cccc|cccc|cccc@{}}
\toprule
\cellcolor[HTML]{FFFFFF}\textbf{Datasets}
& \multicolumn{4}{c|}{Flickr30k}
& \multicolumn{4}{c|}{VQAv2}
& \multicolumn{4}{c}{OKVQA}\\ \midrule

\cellcolor[HTML]{FFFFFF}\textbf{MLLMs}
& \multicolumn{2}{c}{InternVL}
& \multicolumn{2}{c|}{Qwen}
& \multicolumn{2}{c}{InternVL}
& \multicolumn{2}{c|}{Qwen}
& \multicolumn{2}{c}{InternVL}
& \multicolumn{2}{c}{Qwen} \\ \midrule

\rowcolor[HTML]{FFFFFF}
\textbf{Methods}
& RPA$\uparrow$ & $\rho$
& RPA$\uparrow$ & $\rho$
& RPA$\uparrow$ & $\rho$
& RPA$\uparrow$ & $\rho$
& RPA$\uparrow$ & $\rho$
& RPA$\uparrow$ & $\rho$ \\ \midrule

\rowcolor[HTML]{FFFFFF}
KNN
& 0.84 & 0.43
& 0.84 & 0.47
& 0.70 & 0.52
& 0.64 & 0.48
& \underline{0.76} & 0.45
& 0.72 & 0.44 \\

\rowcolor[HTML]{FFFFFF}
LCL
& 0.84 & 0.46
& 0.85 & 0.51
& \textbf{0.80} & 0.59
& \underline{0.70} & 0.55
& \underline{0.76} & 0.47
& 0.71 & 0.36 \\

\rowcolor[HTML]{FFFFFF}
SC-CoT
& \underline{0.90} & 0.39
& \underline{0.89} & 0.37
& \underline{0.76} & 0.56
& 0.68 & 0.50
& 0.70 & 0.51
& 0.71 & 0.42 \\

\rowcolor[HTML]{FFFFFF}
Self-Refine
& 0.86 & 0.29
& 0.87 & 0.25
& 0.55 & 0.32
& 0.62 & 0.33
& 0.75 & 0.40
& \underline{0.81} & 0.41 \\ \midrule

\textbf{COMIL(Ours)}
& \textbf{0.96} & 0.63
& \textbf{0.97} & 0.67
& \underline{0.76} & 0.65
& \textbf{0.73} & 0.68
& \textbf{0.77} & 0.62
& \textbf{0.84} & 0.55 \\ \bottomrule

\end{tabular}%
}
\end{table}

\subsection{Case Study}

\begin{figure}[t]
    \centering
    \includegraphics[width=\columnwidth]{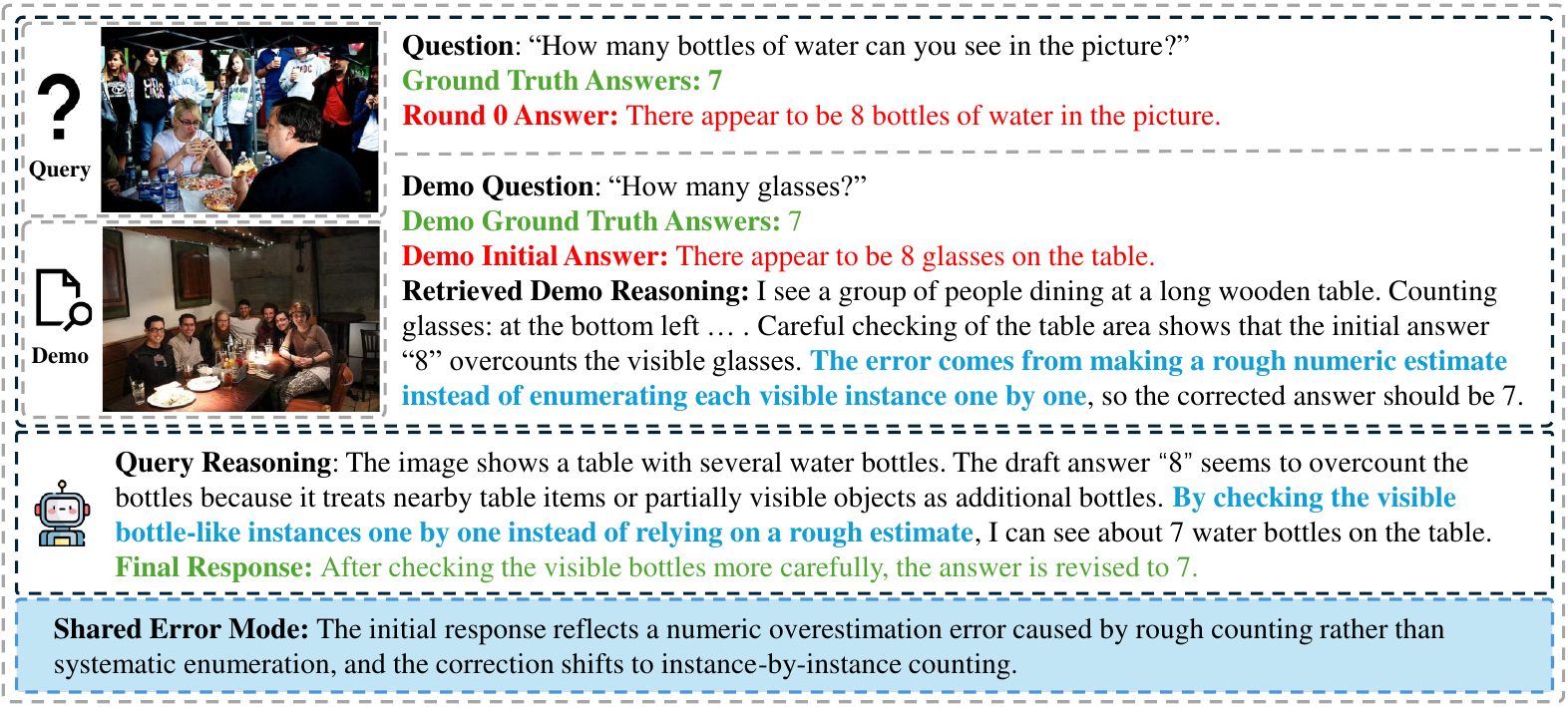}
    \caption{Case study of response-conditioned retrieval.
    Although the query and retrieved demonstration involve different
    objects, both exhibit rough-counting overestimation that is
    corrected by instance-by-instance enumeration.}
    \Description{A VQA case study where both the query and the retrieved demonstration overcount by one; instance-by-instance enumeration provides a transferable correction pattern that changes the response from 8 to 7.}
    \label{fig:retrieval_case}
\end{figure}

As shown in Figure~\ref{fig:retrieval_case}, response-conditioned
retrieval can identify demonstrations with relevant refinement patterns.
In one VQA example, the query asks for the number of water bottles,
for which the initial response overcounts 7 bottles as 8. The retrieved
demonstration asks for the number of glasses, but exhibits the
same error: it predicts 8 instead of 7 because of rough counting. Its
reasoning path corrects this error through instance-by-instance
enumeration, which similarly guides the query response from 8 to 7.
This case illustrates that response-conditioned retrieval can retrieve
transferable correction patterns even when the object semantics differ.
Nevertheless, response similarity serves as a proxy for refinement
relevance rather than a guarantee of the same error mode.

\begin{figure}[t]
    \centering
    \includegraphics[width=\linewidth]{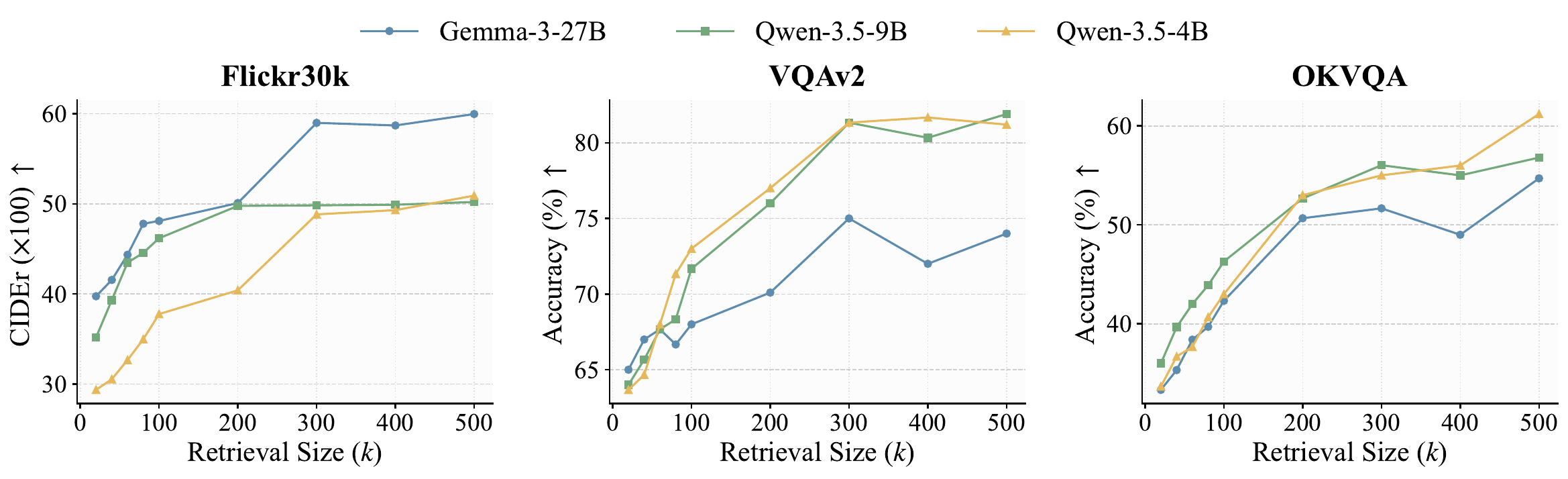}
    \caption{Effect of retrieval set size on COMIL\@. Performance generally improves as the retrieval set grows across Flickr30k, VQAv2, and OKVQA, with diminishing gains at larger scales.}
    \Description{Performance curves for Flickr30k, VQAv2, and OKVQA as the retrieval dataset size increases, showing generally improving performance with diminishing gains at larger retrieval sizes.}
\label{fig:retrieval_size}
\end{figure}

\subsection{Effect of Retrieval Dataset Size} \label{sec:retrieval_dataset_size}

We further study how the size of the contrastive retrieval dataset affects COMIL\@. Specifically, we vary the retrieval dataset size from 20 to 500 and evaluate COMIL on Flickr30k, VQAv2, and OKVQA under three target MLLMs. As shown in Figure~\ref{fig:retrieval_size}, increasing the retrieval dataset size generally improves performance across tasks and MLLMs, suggesting that broader retrieval coverage increases the likelihood of finding demonstrations relevant to the current response. The gains, however, gradually diminish as the retrieval dataset grows. In several settings, performance is already close to its best value at around 300 examples, while further enlargement yields only marginal improvements or small fluctuations. These results indicate that COMIL benefits from richer retrieval coverage but does not require a very large retrieval dataset to achieve strong performance.

\begin{table}[t]
\caption{Results of COMIL on two closed-source MLLMs. COMIL achieves the best performance on CIFAR10 under both Claude Sonnet 4.6 and GPT-4o.}
\label{tab:gpt}
\centering
\resizebox{0.48\textwidth}{!}{%
\begin{tabular}{@{}c|>{\centering\arraybackslash}p{1.8cm}
                >{\centering\arraybackslash}p{1.8cm}
                >{\centering\arraybackslash}p{1.8cm}
                >{\centering\arraybackslash}p{2cm}@{}}
\toprule
\textbf{Methods} & CR & CLIPRE & KNN & \textbf{COMIL(Ours)} \\ \midrule
\textbf{Metrics}  & \multicolumn{4}{c}{Accuracy (\%) $\uparrow$} \\ \midrule
\textbf{MLLMs}   & \multicolumn{4}{c}{Classification: CIFAR10} \\ \midrule
Claude Sonnet 4.6           & 88.1        & 92.5            & 92.9          & 95.3                 \\
GPT-4o           & 96.7        & 98.7            & 99.1          & 99.8                 \\ 
\bottomrule
\end{tabular}
}
\end{table}

\subsection{Extension to Closed-Source MLLMs}

We further evaluate our method on closed-source MLLMs to examine whether its effectiveness extends beyond open-source settings. As shown in Table~\ref{tab:gpt}, our method consistently achieves the best performance on CIFAR10 under both Claude Sonnet 4.6 and GPT-4o~\citep{hurst2024gpt}. Specifically, it improves the accuracy to 95.3\% on Claude Sonnet 4.6 and 99.8\% on GPT-4o, outperforming all retrieval-based baselines in both cases. These results suggest that the benefit of our framework does not depend on a specific open-source architecture, but transfers well to stronger closed-source MLLMs, further supporting the general applicability of the proposed framework.

\subsection{Cost Considerations}

COMIL introduces additional inference cost due to iterative refinement
and is therefore not the lowest-cost ICL method. Nevertheless, compared
with heavier reasoning and refinement baselines, COMIL reduces average
latency and token consumption by 51.6\% and 82.7\%, respectively, over
SC-CoT, and is 29.2\% faster than Self-Refine. The target MLLM remains
frozen, with additional training confined to the lightweight controller.

\section{Conclusion}

In this paper, we present COMIL, a multimodal in-context learning framework that moves beyond surface-level imitation by promoting reasoning path alignment during inference. COMIL reformulates demonstrations as contrastive tuples that describe how a suboptimal response can be refined into a better response, combines them with response-conditioned retrieval, and introduces a lightweight alignment controller to guide refinement. In this way, COMIL enables the MLLM not only to observe better responses in context, but also to refine the current response toward better alignment with the evidence-grounded refinement relation. Extensive experiments on multimodal tasks show that COMIL consistently improves performance, with particularly clear gains on reasoning-intensive tasks. Further reasoning-path analyses and results on closed-source MLLMs support the effectiveness and generality of the proposed framework across different model settings.

\begin{acks}
This work was supported by the National Natural Science Foundation of China
under Grant No. 62502550 and the Shenzhen Science and Technology Program
under Grant No. KJZD20240903095700001.
\end{acks}

\bibliographystyle{ACM-Reference-Format}
\balance
\bibliography{references}

\end{document}